\documentclass[10pt,twocolumn,letterpaper]{article}

\usepackage{cvpr}              
\usepackage{kotex}

\usepackage{amsmath,amssymb} 
\usepackage{multirow}
\usepackage{algorithm}
\usepackage{booktabs}
\usepackage{algpseudocode}
\usepackage{pifont}

\usepackage{lineno}
\usepackage[normalem]{ulem}

\usepackage[dvipsnames]{xcolor}

\definecolor{cvprblue}{rgb}{0.21,0.49,0.74}
\usepackage[pagebackref,breaklinks,colorlinks,citecolor=cvprblue]{hyperref}

\def\paperID{*****} 
\def\confName{CVPR}
\def\confYear{2025}

\title{UniForm: A Reuse Attention Mechanism for Efficient Transformers on Resource-Constrained Edge Devices}

\author{
Seul-Ki Yeom\\
Nota Inc.\\
\\
{\tt\small skyeom@nota.ai}
\and
Tae-Ho Kim\\
Nota Inc.\\
\\
{\tt\small thkim@nota.ai}
}

\begin{document}
\maketitle
\begin{abstract}
Transformer-based architectures have demonstrated remarkable success across various domains, but their deployment on edge devices remains challenging due to high memory and computational demands. In this paper, we introduce a novel Reuse Attention mechanism, tailored for efficient memory access and computational optimization, enabling seamless operation on resource-constrained platforms without compromising performance. Unlike traditional multi-head attention (MHA), which redundantly computes separate attention matrices for each head, Reuse Attention consolidates these computations into a shared attention matrix, significantly reducing memory overhead and computational complexity. Comprehensive experiments on ImageNet-1K and downstream tasks show that the proposed UniForm models leveraging Reuse Attention achieve state-of-the-art imagenet classification accuracy while outperforming existing attention mechanisms, such as Linear Attention and Flash Attention, in inference speed and memory scalability. Notably, UniForm-l achieves a 76.7\% Top-1 accuracy on ImageNet-1K with 21.8ms inference time on edge devices like the Jetson AGX Orin, representing up to a 5x speedup over competing benchmark methods. These results demonstrate the versatility of Reuse Attention across high-performance GPUs and edge platforms, paving the way for broader real-time applications. 
\end{abstract}    
\section{Introduction}
\label{sec:intro}
Recent advances in neural network architectures have been dominated by Transformers, which excel in capturing intricate dependencies across various domains, including natural language processing (NLP), computer vision, and speech recognition. Transformers have proven their worth in large-scale tasks such as image classification, object detection, and language modeling. Vision Transformers (ViTs), in particular, have significantly advanced computer vision, taking inspiration from the breakthroughs seen in large language models (LLMs) such as GPT and Llama. 
However, the impressive performance of these models comes at the cost of increased computational and memory demands, which pose significant challenges for latency-sensitive and real-time applications, particularly on edge devices with limited resources.

\begin{figure}[!t]
\centering
\includegraphics[width=0.99\columnwidth]{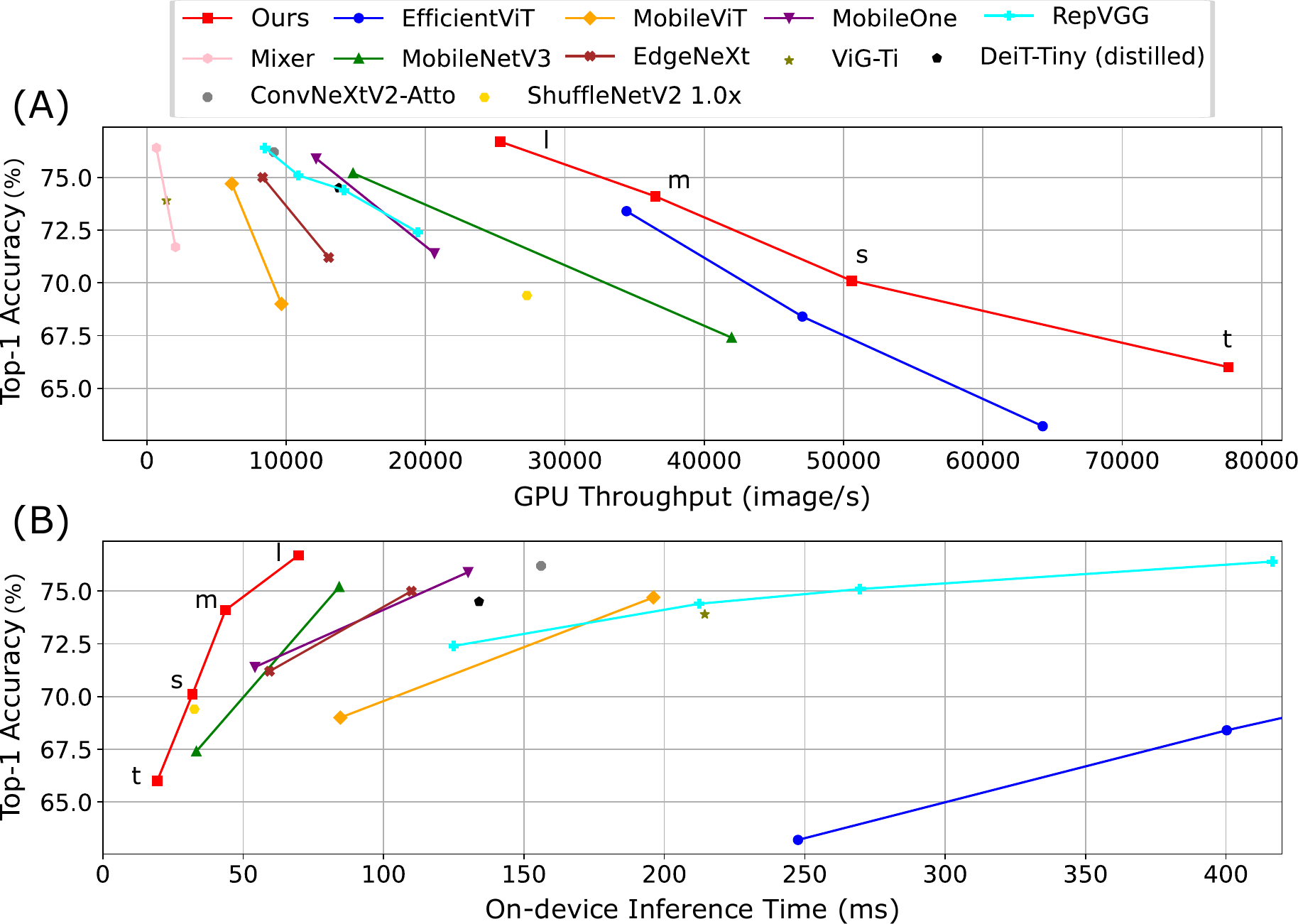}
\caption{Comparison of speed and accuracy between UniForm and other efficient CNN and ViT models, evaluated using the ImageNet-1K dataset in terms of (A) GPU throughput on an NVIDIA A100 and (B) on-device inference time on a Raspberry Pi 4B.}
\label{fig:throughput}
\end{figure}

Edge deployment, such as on devices like Raspberry Pi and Jetson Nano, introduces unique constraints, including limited memory, low energy consumption, and strict real-time processing requirements. The high computational and memory overhead of ViTs and other Transformer-based models makes it difficult to meet these constraints. Even though optimized GPU systems like NVIDIA A100 provide high throughput, the resource disparity between high-end GPUs and edge devices results in a significant performance gap during inference.

To address these challenges, several optimization approaches such as model pruning, quantization, and Neural Architecture Search (NAS) have been explored to optimize models for edge deployment 
However, these techniques often fall short of addressing the non-linear relationship between model complexity and inference time on resource-constrained platforms. This non-linearity has been well-documented in studies such as MCUNet and Neural Architecture Search for Efficient Edge Inference, which highlight that models designed for GPUs often fail to scale efficiently to edge environments~\cite{JianRen23,DettmersZ23}. This gap emphasizes the need for models designed specifically for edge hardware. 

In response to these challenges, we propose a simple yet efficient reuse attention mechanism that is specifically designed to address the constraints of edge devices while maintaining the scalability and accuracy of ViTs. Unlike traditional attention mechanisms, which calculate the attention matrix for each head, our method reuses the attention matrix across multiple attention heads within a layer. This significantly reduces redundant computations and minimizes memory accesses, a crucial improvement for memory-bound edge devices. This approach is particularly advantageous in real-time, latency-sensitive applications, offering higher inference speeds and reduced memory usage. Despite the optimizations, our reuse attention mechanism can also maintain competitive performance on tasks like object detection and segmentation, ensuring its applicability across different downstream tasks.


In summary, our contributions are threefold:

\begin{itemize}
    \item \textit{Efficient Attention Mechanism} We propose a simple yet efficient attention mechanism that calculates the attention matrix once per layer, significantly reducing redundant calculations and memory access. This innovation enables more efficient processing on memory-constrained edge devices while maintaining computational rigor, thus addressing the growing demand for real-time, low-latency AI applications.
    \item \textit{Improved Throughput on Diverse Hardware} Our method demonstrates significant improvements in throughput across diverse hardwares, ranging from high-performance GPUs to resource-constrained edge devices. Unlike conventional attention mechanisms, our approach scales gracefully without compromising on Top-1 accuracy, outperforming many state-of-the-art (SOTA) models.
    \item \textit{Versatility in Downstream Tasks} We extend the applicability of Reuse Attention to downstream tasks like object detection and segmentation. This versatility shows its practical utility in real-world applications while maintaining computational efficiency.
\end{itemize}

\section{Backgrounds}
\label{sec:background}
Deploying Transformers on edge devices presents unique challenges due to the limited computational and memory resources available in these environments. This section highlights three major bottlenecks: (1) memory bottlenecks from multi-head attention (MHA), (2) computational inefficiencies and redundancy in traditional attention modules, and (3) the performance gap between high-performance GPUs and resource-constrained edge devices. These challenges have a significant impact on the scalability and real-time deployment of ViTs, underscoring the need for optimized transformer architectures that can effectively operate in low-resource environments while maintaining performance.

\subsection{Memory Bottlenecks}
Attention mechanisms, particularly multi-head attention (MHA), are foundational to the success of Transformer-based models, including Vision Transformers (ViTs) as well as Large Language Models, enabling them to capture long-range dependencies and global relationships in input data~\cite{VaswaniNeurIPS17Attention}. However, MHA introduces significant memory bottlenecks, especially in resource-constrained environments like edge devices~\cite{LiICCV23}. The core issue lies in the memory-intensive operations involved in attention modules, such as softmax calculations and matrix multiplications, which require frequent memory access to intermediate results and output storage, placing high demands on memory bandwidth~\cite{KongECCV22SPViT}.
The frequent data movement between memory and compute units significantly increases memory traffic as the input sequence size grows, which leads to slower inference times for real-time applications.

High-performance GPUs, such as the NVIDIA H100 with 3.35 TB/s bandwidth, are well-equipped to handle the memory demands of transformer models, particularly multi-head attention (MHA). In contrast, edge devices like the Raspberry Pi 3B, with about 197 times less bandwidth (17 GB/s), struggle significantly as shown in Fig.~\ref{fig:device_spec}. The low memory bandwidth in edge devices forces frequent memory I/O operations, increasing memory traffic and leading to performance bottlenecks like cache misses, especially during repetitive MHA operations. This negatively impacts inference speed as the processor repeatedly accesses slower memory~\cite{LanWACV23Couplformer, KongECCV22SPViT}. Previous studies have shown that these memory inefficiencies can be mitigated through optimizations such as FlashAttention, which reduces unnecessary memory transfers, minimizing latency and improving throughput on low-resource hardware~\cite{DongICML23, DaoNeurIPS22Flash} 

\subsection{Computation and Parameter Optimization}
Computational inefficiencies in traditional MHA modules, particularly in resource-constrained environments, present a significant challenge. Each attention head computes its attention map independently, leading to redundant and resource-intensive operations as the number of heads increases. Studies have shown that many attention maps across different heads exhibit substantial similarity, suggesting that not all computations are necessary~\cite{LiuCVPR23EfficientViT}. Furthermore, Michel et al.~\cite{Michelneurips19} demonstrate that transformer models often rely on only a subset of heads, with many contributing minimally to the final output. This underscores the diminishing returns of additional heads and highlights the potential benefits of simplified architectures, such as single-head attention or reduced multi-head configurations, which strike a better balance between computational efficiency and performance.

Moreover, parameter efficiency is equally critical, particularly for edge device deployments. Research on ViT pruning~\cite{YangCVPR23} reveals redundancy in the Query and Key components of the attention module across layers, while the Value component consistently retains essential information. This finding suggests that selectively pruning or simplifying the Query and Key components while preserving the Value can significantly reduce parameter overhead. Such optimizations lead to improved memory and computational efficiency without compromising performance, enabling more practical and efficient transformer designs for deployment in low-power environments.

\subsection{Performance Gap between High-performance GPUs vs. Edge Devices}
The increasing complexity of Transformer-based models, such as Vision Transformers (ViTs), highlights the growing need for more efficient mechanisms, especially when deploying models across a variety of hardware platforms. A distinct performance gap exists between high-performance GPUs and Edge-devices, primarily due to the computational and memory demands of conventional attention mechanisms. While high-performance GPUs can manage these demands, Edge-devices face significant challenges due to limited computational power and memory bandwidth, making efficient deployment difficult.

\begin{figure}[t]
\centering
\includegraphics[width=0.95\columnwidth]{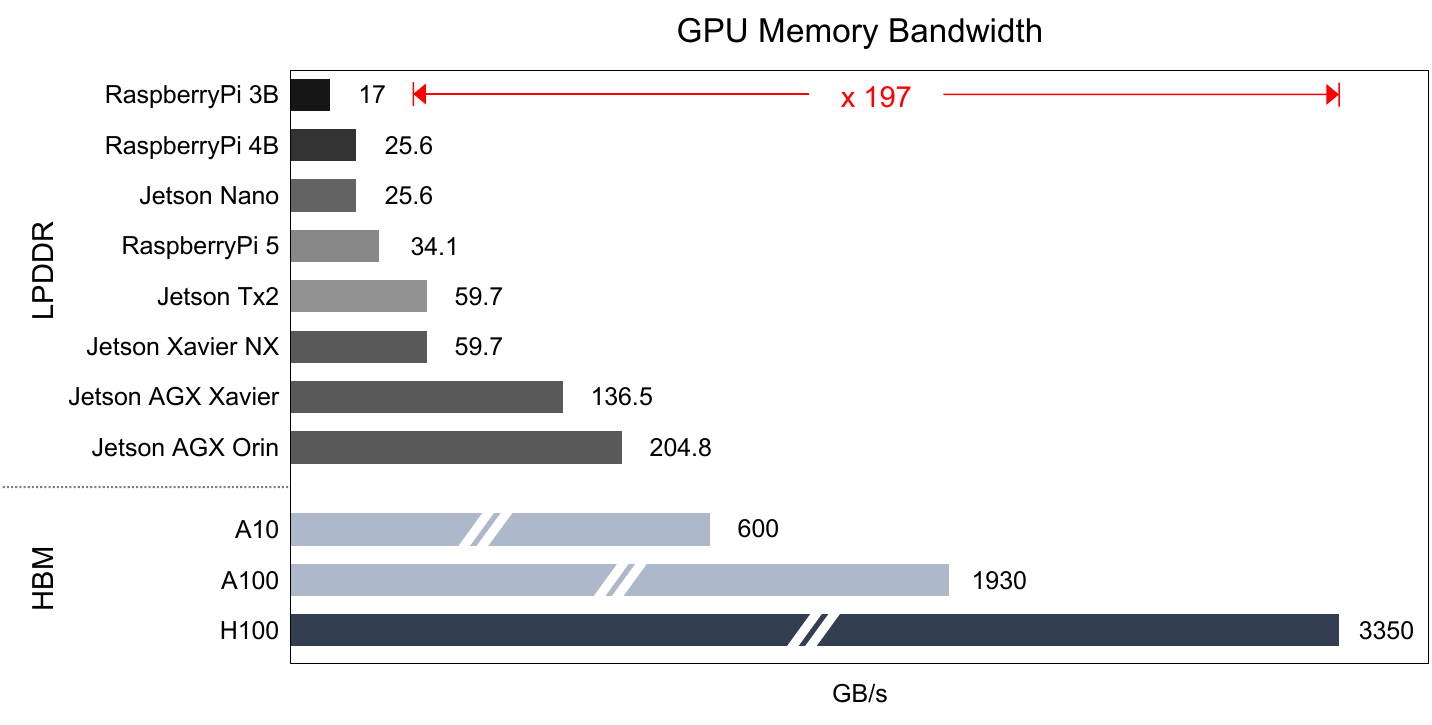}
\caption{Comparison of Memory Bandwidth between High-Performance GPUs and Edge Devices}
\label{fig:device_spec}
\end{figure}

High-performance GPUs, equipped with High Bandwidth Memory (HBM), can efficiently handle these demands, processing large sequences with minimal latency. For example, the H100 offers a memory bandwidth of 3.35 TB/s, which is approximately 197 times greater than the Raspberry Pi 3B, allowing it to smoothly manage increasing token counts. In contrast, Edge-devices, such as the Jetson Nano with LPDDR memory (25.6 GB/s), are heavily constrained by lower memory bandwidth and compute capacity (see Figure~\ref{fig:device_spec}). As the number of tokens increases, memory access in conventional attention mechanisms grows exponentially, overwhelming these devices. This results in frequent cache misses, increased latency, and overall inefficiency when handling multi-head attention (MHA) operations, which require frequent memory input/output (I/O) actions.

\begin{figure}[t]
\centering
\includegraphics[width=0.95\columnwidth]{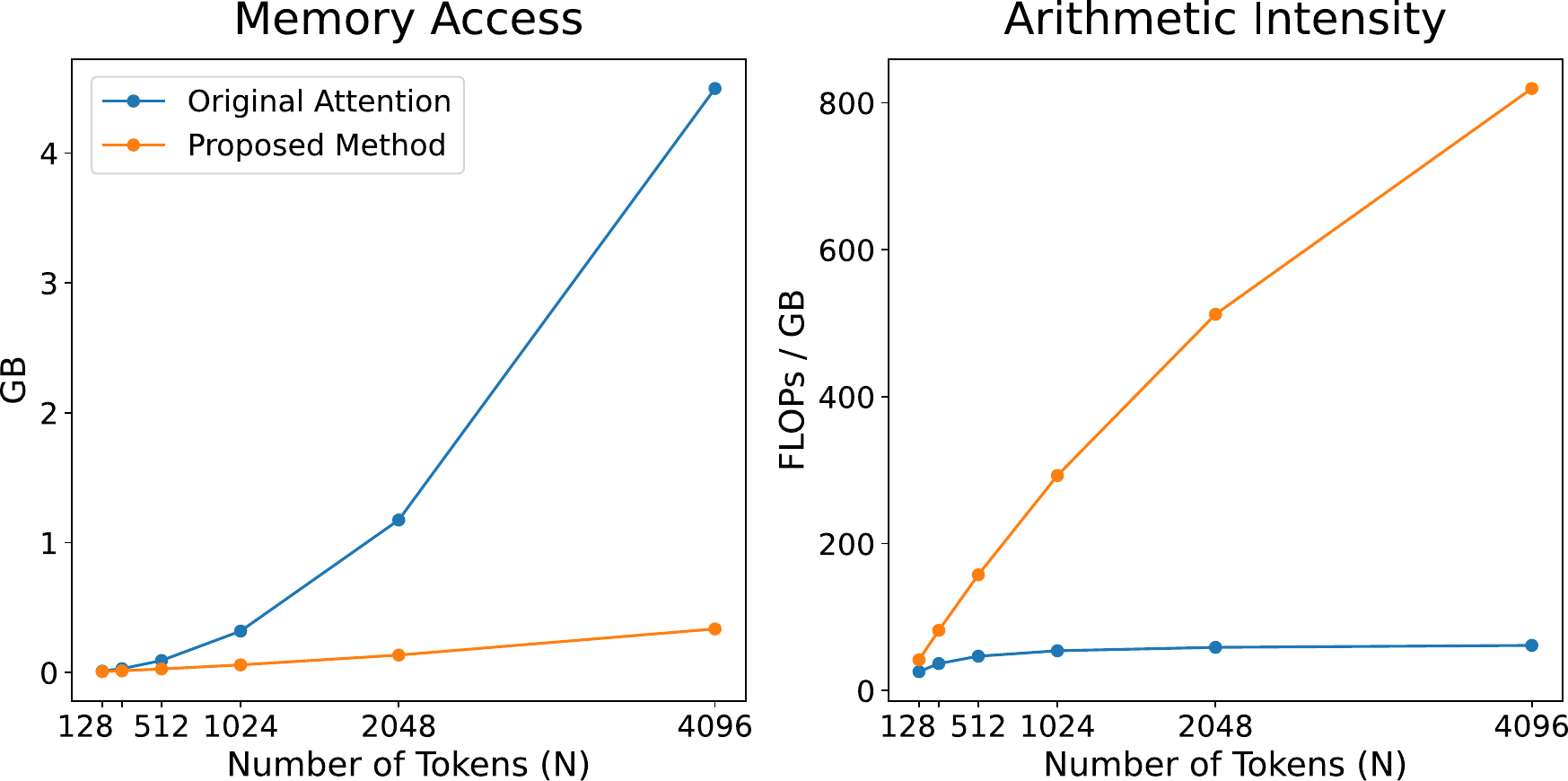}
\caption{Impact of the number of tokens on Memory Access and Arithmetic Intensity in Attention.}
\label{fig:attention_compare}
\end{figure}

Our experiments clearly validated this performance gap as shown in Figure~\ref{fig:attention_compare}. By evaluating the memory access and arithmetic intensity of attention modules across devices—from high-end GPUs such as the H100 and A100 to Edge-devices like the Jetson Nano and Raspberry Pi—we observed that when token counts reached 1024 or higher, memory access became the primary bottleneck for Edge-devices. While high-performance GPUs continued to perform well under these conditions, Edge-devices experienced significant slowdowns due to their limited resources. Additionally, the increased number of memory accesses led to frequent cache misses, further degrading performance, underscoring the need for specialized optimizations in attention mechanisms. Without addressing these memory bandwidth constraints, deploying Transformer-based models on Edge-devices for real-time or large-scale input processing remains a major challenge.

\begin{table}[]
    \centering
    \caption{Memory movement comparison between Multi-Head and Reuse Attention.}
    \label{tab:attention_comparison11}
    \newcommand{\spheading}[3][6em]{
    \rotatebox{90}{\parbox{#1}{\raggedright #2}}}
    \scalebox{0.60}{
    \setlength{\tabcolsep}{2pt}
\begin{tabular}{ccccc}
\toprule
  & \multicolumn{2}{c}{\textbf{Multi-Head Attention}} & \multicolumn{2}{c}{\textbf{Reuse Attention}} \\
    & Memory Load & Memory Store    & Memory Load & Memory Store \\
\midrule
\boldmath{$S = QK^T$} &  $2 \times (N \times d) \times h$                &   $N^2 \times h$    &   $2 \times (N \times D)$                &    $N^2$             \\
\boldmath{$P = Softmax(S)$}   &    $N^2 \times h$                 &   $N^2 \times h$     &      $N^2$     &    $N^2$    \\
\boldmath{$O=PV$}        &   $(N^2 + (N \times d)) \times h$     &    $(N \times d) \times h$    &    $N^2 + (N \times d \times h)$     &      $(N \times d) \times h$          \\
\bottomrule
\end{tabular}
    }
\end{table}

\begin{table}[]
    \centering
    \caption{Memory movement comparison between multi-head and reuse Attention across various model categories, including LLM (Llama 2 7B, GPT-3, ALBERT xxlarge, BERT Large, Transformer XL, T5 Large), VLM (CLIP Text Encoder), ViT (ViT Base), and Text-to-Image Generation (Stable Diffusion). Reuse Attention demonstrates a significant reduction in memory movement compared to multi-head attention.
}

    \label{tab:attention_comparison11}
    \newcommand{\spheading}[3][6em]{
    \rotatebox{90}{\parbox{#1}{\raggedright #2}}}
    \scalebox{0.70}{
    \setlength{\tabcolsep}{2pt}
\begin{tabular}{lccc}
\toprule
\textbf{Model}            & \textbf{Multi-Head Attention} & \textbf{Reuse Attention} & \textbf{Reduction (\%)} \\
\midrule
\textbf{Llama 2 7B}                & 141.73 GB                       & 8.59 GB                      & 93.94                   \\
\textbf{GPT-3}                     & 328.56 GB                       & 22.55 GB                      & 93.14                   \\
\textbf{ALBERT xxlarge}            & 1.81 GB                          & 226.49 MB                       & 87.50                   \\
\textbf{BERT Large}                & 905.97 MB                          & 150.99 MB                        & 83.33                   \\
\textbf{Transformer XL}            & 679.48 MB                          & 113.25 MB                        & 83.33                   \\
\textbf{T5 Large}                  & 905.97 MB                          & 150.99 MB                        & 83.33                   \\
\textbf{CLIP Text Encoder}         & 8.34 MB                             & 4.35 MB                          & 47.78                   \\
\textbf{ViT Base}                  & 59.23 MB                           & 18.25 MB                          & 69.19                   \\
\textbf{Stable Diffusion}          & 16.68 MB                           & 8.33 MB                          & 50.06                   \\
\bottomrule
\end{tabular}
    }
\end{table}

\begin{table}[]
    \centering
    \caption{Comparison of attention mechanisms in terms of key performance metrics: model performance, GPU efficiency, edge device efficiency, and memory scalability. Each category is rated on a scale of one to three stars (\ding{72}), with more stars indicating superior performance in that category.}
    \label{tab:attention_comparison}
    \newcommand{\spheading}[3][6em]{
    \rotatebox{90}{\parbox{#1}{\raggedright #2}}}
    \scalebox{0.69}{
    \setlength{\tabcolsep}{6pt}
\begin{tabular}{ccccccc}
\toprule
      & Model & GPU & Edge & Memory \\
      & Performance & Efficiency & Efficiency & Scalability \\
\midrule
Attention          &     \ding{72} \ding{72}              &      \ding{72}         &         \ding{72}        &     \ding{72}               \\
Linear Attention   &     \ding{72}              &            \ding{72} \ding{72} \ding{72}    &         \ding{72} \ding{72}        &       \ding{72} \ding{72}             \\
Cascaded Attention &      \ding{72} \ding{72} \ding{72}             &      \ding{72} \ding{72} \ding{72}          &          \ding{72}       &        \ding{72}            \\
Flash Attention    &  \ding{72} \ding{72}                 &      \ding{72} \ding{72} \ding{72}          &     \ding{72}            &         \ding{72}           \\
\textbf{Ours}               &   \ding{72} \ding{72} \ding{72}                &      \ding{72} \ding{72} \ding{72}            &         \ding{72} \ding{72} \ding{72}          &     \ding{72} \ding{72} \ding{72}       \\
\bottomrule
\end{tabular}
    }
\end{table}

\subsection{Related Works: Comparative Analysis of Attention Mechanisms}
Building on the identified three primary bottlenecks, we evaluate popular attention mechanisms (i.e., Original Attention, Linear Attention, Cascaded Group Attention, and Flash Attention) alongside our proposed Reuse Attention approach, focusing on four key metrics: model performance, GPU efficiency, edge efficiency, and memory scalability (see Table~\ref{tab:attention_comparison}).

Conventional attention mechanisms, though delivering robust model performance, impose substantial computational and memory demands. These demands scale sharply with increased token counts, model depth, and the number of heads, ultimately constraining efficiency on both GPUs and edge devices. This limitation has been a well-documented obstacle in scaling Transformer models to edge environments~\cite{VaswaniNeurIPS17Attention, ShenWACV21, DaoNeurIPS22Flash}.

Linear Attention improves GPU and edge efficiency by reducing computational complexity. However, this simplification often sacrifices accuracy, as it struggles to capture complex dependencies required for tasks like language modeling. Consequently, Linear Attention may be less suitable for high-precision applications where performance is critical~\cite{KatharopoulosICML20LinearAttn, SchlagICML21, HanICCV23, LuNeurIPS21}.

Cascaded Group Attention achieves high accuracy, but its grouped computations reduce efficiency on edge devices, where limited parallelism and bandwidth lead to significant performance drops. Prior work has highlighted the resource constraints that prevent grouped attention mechanisms from reaching optimal efficiency on hardware-limited platforms~\cite{LiuCVPR23EfficientViT}.

Flash Attention enhances GPU efficiency by employing block-wise computation to reduce memory I/O, but it still encounters training instability and demands significant memory bandwidth, particularly at higher embedding dimensions, which limits its practicality on SRAM-limited edge devices~\cite{DaoNeurIPS22Flash, Goldenarxiv24}.

In contrast, our proposed Reuse Attention significantly reduces memory and computational demands by reusing the attention matrix across heads and employing multi-scale value processing. This design achieves high accuracy while enhancing efficiency on both GPUs and edge devices, making it a viable option for real-world edge deployments. Preliminary results show that Reuse Attention retains accuracy comparable to conventional attention but with a reduced resource footprint, aligning with findings in recent efforts to optimize attention mechanisms for constrained environments.
\section{Proposed Method}
\label{sec:proposed_method}
In this section, we introduce the Reuse Attention Mechanism with Multi-Scale Value Processing, a simple yet efficient approach designed to mitigate the computational and memory bottlenecks of the conventional Multi-Head Attention (MHA) mechanism in Transformers. Our method focuses on reducing redundant computations and minimizing memory access overhead, which are critical factors affecting inference speed and efficiency on edge devices. By reusing the attention matrix across all heads and incorporating multi-scale processing in the value projections, our approach enhances both computational efficiency and representational diversity. 

\subsection{Redundancy in Query and Key Components}
Studies have observed that attention maps exhibit high similarities across different heads, leading to computational redundancy~\cite{LiuCVPR23EfficientViT}. This redundancy suggests that computing separate attention matrices for each head may be unnecessary. Additionally, Mehta et al.  demonstrated that using synthetic or fixed attention patterns can maintain or even improve performance, aligning with our motivation to reuse the attention matrix~\cite{MehtaICLR21}.

\subsection{Importance of Value Projections}
Recent research exploring the pruning of various structural components of Vision Transformers indicates that reducing the dimensionality of the Value projections leads to greater performance degradation compared to reducing the dimensions of the Query and Key projections~\cite{YangCVPR23}. This suggests that the Value projections potentially encode more crucial information for the model's performance. Therefore, we maintain the original dimensionality of the Value projections while reusing the attention matrix to enhance efficiency without compromising expressiveness.

\subsection{Enhancing Representational Diversity with Multi-Scale Processing} 
Our multi-scale Value processing is inspired by the effectiveness of multi-scale convolutional strategies, such as MixConv~\cite{TanBMVC19} and Inception modules~\cite{SzegedyCVPR15}. Each Value projection undergoes depthwise convolution with a unique kernel size, enabling each head to capture features at different receptive fields. This design improves the model's ability to learn rich and diverse representations essential for complex tasks, particularly in edge environments where efficiency and flexibility are crucial.

\begin{figure*}[t]
\centering
\includegraphics[width=1.7\columnwidth]{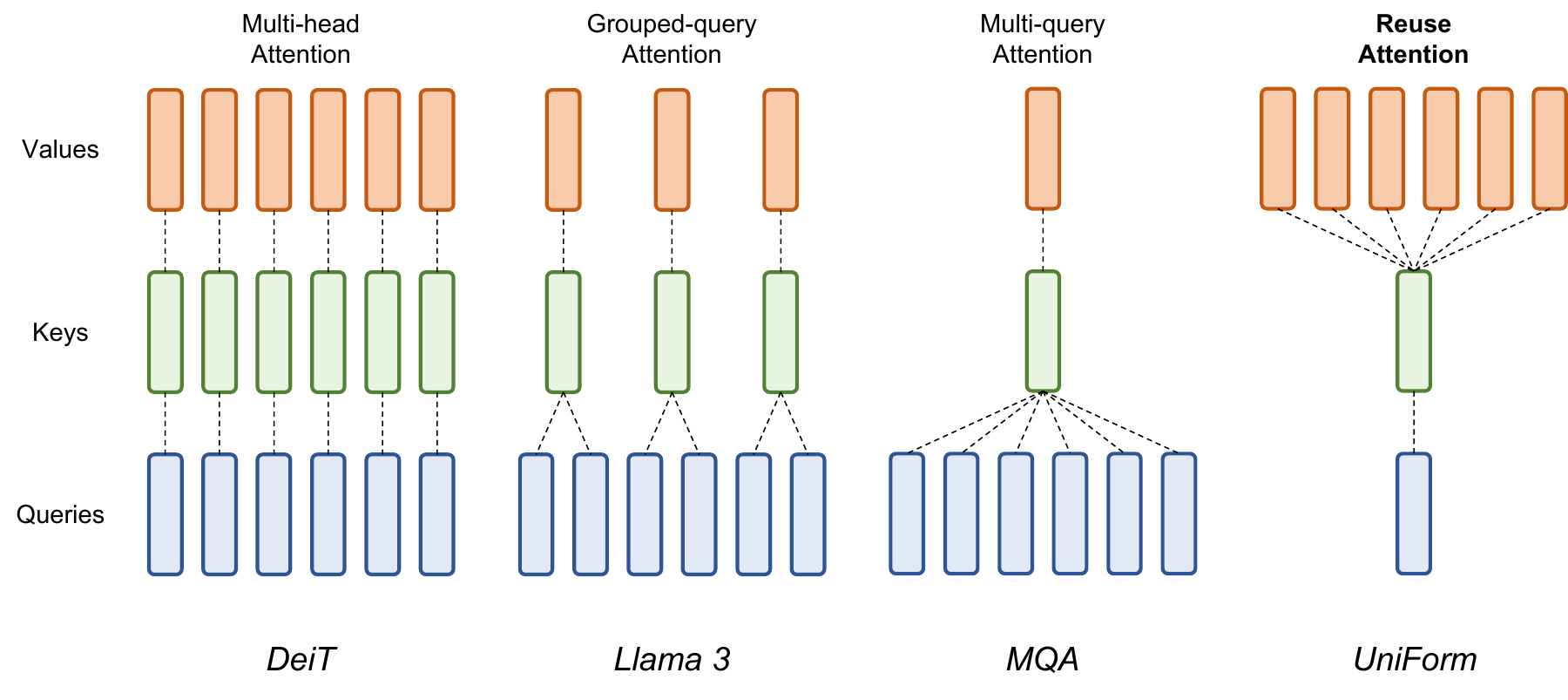}
\caption{Overview of the Proposed Method compared to the previous attention mechanisms}
\label{fig:overview}
\end{figure*}

\subsection{Reuse Attention: Overview and Computational Steps}

Our Reuse Attention Mechanism with Multi-Scale Value Processing is a streamlined approach designed to address the inefficiencies in conventional attention architectures. By reusing a single attention matrix across all heads and incorporating multi-scale processing within Value projections, our method reduces computational redundancy and minimizes memory overhead, enhancing efficiency on edge devices
\paragraph{Comparative Overview of Attention Mechanisms} 
Figure~\ref{fig:overview} illustrates the architectural differences between prominent attention mechanisms, highlighting how Queries, Keys, and Values are handled across architectures compared to our proposed method.

\begin{itemize}
    \item \textbf{Multi-Head Attention (DeiT)}: 
    This configuration employs independent Query, Key, and Value projections for each head, which allows diverse representations but introduces high memory and computational demands, especially with increased heads or token counts. Recent studies highlight that the growing number of heads in Multi-Head Attention exacerbates memory and compute constraints, limiting its scalability~\cite{ChenNeurIPS24}
    
    \item \textbf{Grouped-Query Attention (Llama 3)}: 
    In this approach, Queries are grouped, with each group sharing a representative Key-Value pair. While this setup reduces resource requirements, it may sacrifice specificity within groups, as each Key-Value pair must represent a range of Queries. This trade-off has been explored as a means to improve efficiency without entirely sacrificing representational power, although some group configurations can hinder task-specific performance~\cite{AinslieEMNLP23}.
    
    \item \textbf{Multi-Query Attention}: This method goes further by consolidating queries with a single Key-Value pair to improve efficiency, but its capacity to capture diverse structures is restricted. These methods face inherent trade-offs between performance and efficiency, and they lack scalability, especially in edge environments with limited resources, particularly when dealing with high-dimensional embeddings~\cite{Brandonarxiv24}.
\end{itemize}

\paragraph{Reuse Attention with Multi-Scale Value Processing}
To address the limitations of existing attention mechanisms, we propose \textit{Reuse Attention} with Multi-Scale Value Processing, which reduces redundant computations and minimizes memory access overhead while preserving the expressive power of MHA. Our Reuse Attention mechanism leverages a unified attention matrix shared across all heads and introduces multi-scale processing in the value projections, thereby maintaining efficiency and enhancing the model’s representational capacity, especially for edge deployment.

Our Reuse Attention mechanism maximizes computational efficiency by creating a single attention matrix shared across all heads. Given an input $X \in \mathbb{R}^{N \times D}$, we compute the shared Query and Key projections as follows:

\begin{equation} 
Q = X W^Q, \quad K = X W^K, 
\end{equation}

The shared attention matrix $A$ is then calculated using:

\begin{equation} 
A = \text{softmax}\left( \frac{Q K^\top}{\sqrt{D}} \right), 
\end{equation}

where $A \in \mathbb{R}^{N \times N}$ captures the global attention across the entire dimension.

Though the matrix multiplication complexity remains $O(N^2 D)$, the proposed method significantly reduces memory I/O operations by eliminating the need to separately compute attention matrices for each head. In traditional multi-head attention, each head computes its own attention matrix, resulting in redundant memory accesses and increased I/O overhead. By reusing a single attention matrix $A$ across all heads, our method reduces the total amount of data that needs to be read from and written to memory, thereby improving memory efficiency.

\begin{figure*}[t]
\centering
\includegraphics[width=1.7\columnwidth]{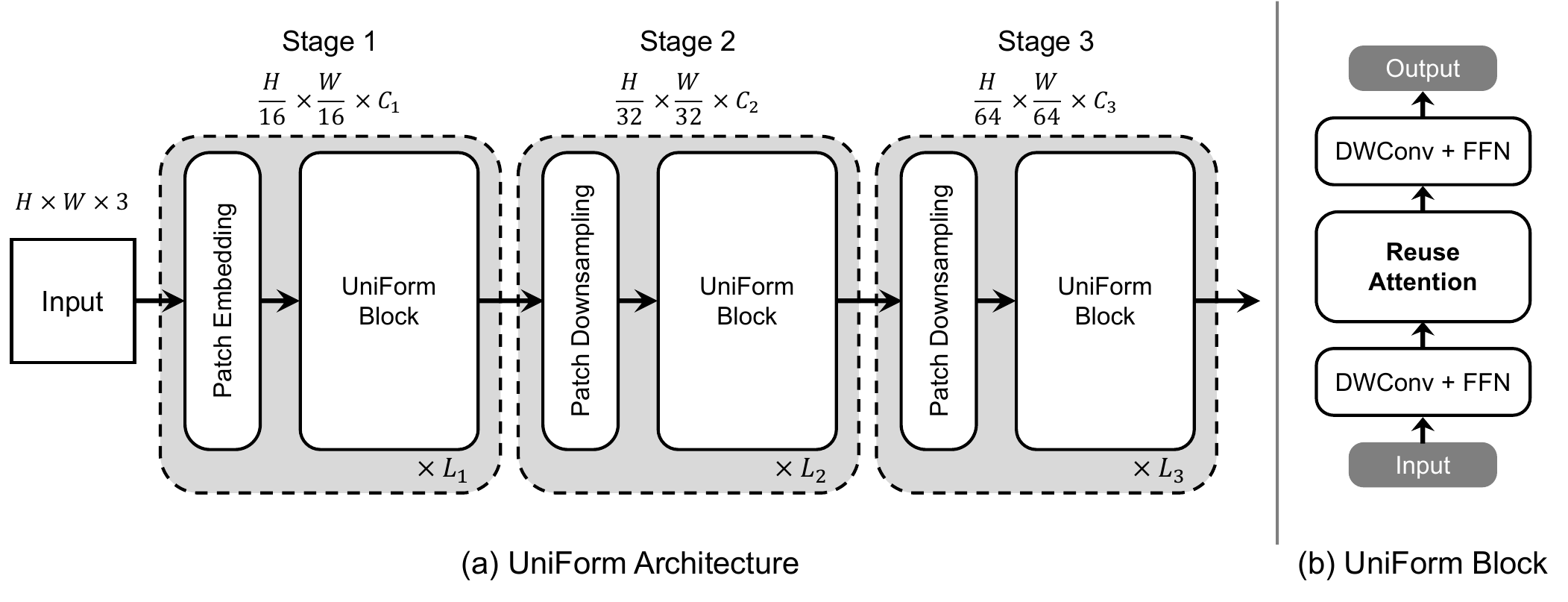}
\caption{(a) The architecture of UniForm; (b) UniForm Block including Reuse Attention.}
\label{fig:architecture}
\end{figure*}

Recent research demonstrated that minimizing memory transfers between compute units and memory banks can drastically improve efficiency on memory-bound hardware~\cite{DaoNeurIPS22Flash}. By reusing $A$ across all heads, our approach parallels the memory efficiency strategies of Flash Attention, which minimizes bandwidth requirements through reduced memory transfers, effectively lowering memory traffic, and computational overhead per head. This design addresses a key limitation in transformer architectures by significantly decreasing memory I/O without increasing computational complexity, thereby enhancing throughput on memory-constrained devices~\cite{PanNeurIPS22, LiuCVPR23EfficientViT, ZhengICCVw23}. 

Each head in the Reuse Attention Mechanism processes its Value projection using depthwise convolutions with unique kernel sizes, enabling multi-scale feature extraction:

\begin{equation} V_h = \text{DWCONV}_{k_h}(X_h W^{V_h}), \end{equation}

By applying depthwise convolutions with varying kernel sizes $k_h$ across heads, our method captures multi-scale contextual information. Smaller kernels focus on fine-grained details, while larger kernels encompass broader context. This approach allows each head to extract features at different scales without increasing the memory I/O, as the convolution operations are performed locally within the Value projections. Inspired by studies~\cite{TanBMVC19, YangCVPR23}, which demonstrated the effectiveness of multi-scale convolutions in capturing spatial hierarchies, our approach further improves representational diversity across scales. This multi-scale processing enables a richer representation without requiring additional memory bandwidth, aligning well with insights on optimizing model expressiveness through hierarchical representation without inflating parameter size or memory requirements.

By reusing the unified attention matrix $A$ across all heads, the Reuse Attention Mechanism achieves a significant reduction in redundant memory operations:

\begin{equation} \textit{O}_{h} = A V_h, \end{equation}

Unlike conventional attention, which calculates separate attention matrices per head, our method reuses the same $A$ for all heads, thus avoiding repeated memory access patterns and reducing the total memory bandwidth required. This reuse strategy is akin to techniques in recent research like that of Ribar et al.~\cite{RibarICML24SparQ} in LLMs and Shim et al.~\cite{Shimarxiv23reuse} in speech recognition, which illustrate how minimizing repeated memory accesses can lead to substantial performance gains in constrained environments. Our approach ensures that the high memory bandwidth demand seen in conventional MHA is alleviated, aligning with recent findings on optimizing memory efficiency in transformers.

Finally, the outputs from each head are concatenated to form the final output:

\begin{equation} \text{Output} = \text{Concat}\left( \textit{O}_1, \textit{O}_2, \dots, \textit{O}_H \right), \end{equation}

This aggregation leverages the multi-scale, memory-efficient representations from each head, preserving high expressiveness without overburdening memory resources. By reducing memory transfers and managing bandwidth effectively, our method provides significant improvements in memory efficiency, which is crucial for edge applications with limited memory resources. This aligns well with insights from existing studies on transformer efficiency for edge applications.

\begin{table}[]
    \centering
    \caption{Architecture detail of UniForm model variants.}
    \label{tab:model_architecture}
    \scalebox{0.8}{
    \setlength{\tabcolsep}{6pt}
\begin{tabular}{cccc}
\toprule
\multirow{2}{*}{\textbf{Architecture}}     & \textbf{Channel}  & \textbf{Depth} & \textbf{Head}  \\
& $[C_1, C_2, C_3]$ & $[L_1, L_2, L_3]$ & $[H_1, H_2, H_3]$ \\
\midrule
Tiny & $[64, 128, 192]$                                                        & $[1, 2, 2]$                                                          & $[4, 4, 4]$                                                          \\
Small & $[128, 144, 192]$                                                       & $[1, 2, 2]$                                                           & $[4, 4, 4]$                                                          \\
Medium & $[128, 240, 320]$                                                       & $[1, 2, 3]$                                                           & $[4, 3, 4]$                                                          \\
Large & $[192, 288, 384]$                                                       & $[1, 2, 3]$                                                           & $[3, 3, 4]$                                             \\
\bottomrule
\end{tabular}
}
\end{table}

\begin{table*}[t]
    \centering
    \caption{Performance comparison with the state-of-the art CNN and ViT models on ImageNet-1K.}
    \label{tab:perform_comparison_acc}
    \scalebox{0.70}{
    \setlength{\tabcolsep}{6pt}
\begin{tabular}{cccccccc}
\toprule
\multirow{2}{*}{\textbf{Model}} & \textbf{Base} & \textbf{Top-1 acc.} & \textbf{Top-5 acc.} & \multicolumn{2}{c}{\textbf{Throughput (images/s)}} & \textbf{FLOPs} & \textbf{Params} \\
                       &     \textbf{architecture}                               &  ($\%$)~$\uparrow$                         & ($\%$)~$\uparrow$                            & \textbf{GPU}~$\uparrow$                & \textbf{CPU}~$\uparrow$                &  \textbf{(M)}~$\downarrow$                          & \textbf{(M)}~$\downarrow$                            \\
\midrule
EfficientViT-M0        & Transformer                        & 63.2                        & 85.4                        & 64293                & 450                & 79                         & 2.3                         \\
\textbf{UniForm-t}     & \textbf{Transformer}               & \textbf{66.0}               & \textbf{86.6}               & \textbf{77625}       & \textbf{544}       & \textbf{74}                & \textbf{1.8}                \\
\midrule
MobileNetV3-small      & CNN                                & 67.4                        & 87.4                        & 41965                & 360                & 57                         & 2.5                         \\
EfficientViT-M1        & Transformer                        & 68.4                        & 88.7                        & 47045                & 220                & 167                        & 3.0                         \\
MobileViT-XXS          & Transformer                        & 69.0                        & 88.9                        & 9663                 & 59                 & 410                        & 1.3                         \\
ShuffleNetV2 1.0x      & CNN                                & 69.4                        & 88.9                        & 27277                & 138                & 146                        & 2.3                         \\
\textbf{UniForm-s}     & \textbf{Transformer}               & \textbf{70.1}               & \textbf{89.3}               & \textbf{50582}       & \textbf{231}       & \textbf{164}               & \textbf{2.4}                \\
\midrule
EdgeNeXt-XXS           & Both                               & 71.2                        & -                           & 13051                & 121                & 261                        & 1.3                         \\
MobileOne-S0           & CNN                                & 71.4                        & 89.8                        & 20642                & 26                 & 275                        & 2.1                         \\
Mixer-B/16             & MLP                                & 71.7                        & -                           & 2057                 & 6                  & 12610                      & 59.8                        \\
RepVGG-A0              & CNN                                & 72.4                        & -                           & 19450                & 61                 & 1366                       & 8.3                         \\
EfficientViT-M3        & Transformer                        & 73.4                        & 91.4                        & 34427                & 166                & 263                        & 6.9                         \\
ViG-Ti                 & GNN                                & 73.9                        & 92.0                        & 1406                 & 6                  & 1300                       & 7.1                         \\
\textbf{UniForm-m}     & \textbf{Transformer}               & \textbf{74.1}               & \textbf{91.9}               & \textbf{36507}       & \textbf{174}       & \textbf{251}               & \textbf{5.6}                \\
\midrule
RepVGG-A1              & CNN                                & 74.4                        & -                           & 14155                & 39                 & 2362                       & 12.7                        \\
DeiT-Tiny (distilled)  & Transformer                        & 74.5                        & -                           & 13785                & 63                 & 1085                       & 5.9                         \\
MobileViT-XS           & Transformer                        & 74.7                        & 92.3                        & 6098                 & 13                 & 986                        & 2.3                         \\
ShuffleNetV2 2.0x      & CNN                                & 74.9                        & 92.4                        & 12910                & 67                 & 591                        & 7.4                         \\
EdgeNeXt-XS            & Both                               & 75.0                        & -                           & 8312                 & 69                 & 538                        & 2.3                         \\
RepVGG-B0              & CNN                                & 75.1                        & -                           & 10868                & 30                 & 15824                      & 14.3                        \\
MobileNetV3-large      & CNN                                & 75.2                        & 91.3                        & 14798                & 69                 & 217                        & 5.4                         \\
MobileOne-S1           & CNN                                & 75.9                        & 92.5                        & 12150                & 22                 & 825                        & 4.8                         \\
ConvNeXtV2-Atto        & CNN                                & 76.2                        & 93.0                        & 9120                 & 73                 & 552                        & 3.7                         \\
Mixer-L/16             & MLP                                & 76.4                        & -                           & 688                  & 2                  & 44570                      & 208.2                       \\
RepVGG-A2              & CNN                                & 76.4                        & -                           & 8483                 & 20                 & 5123                       & 25.4                        \\
\textbf{UniForm-l}     & \textbf{Transformer}               & \textbf{76.7}               & \textbf{93.2}               & \textbf{25356}       & \textbf{113}       & \textbf{467}               & \textbf{10.0}           \\           
 \bottomrule
\hline
\end{tabular}
}
\end{table*}

\subsection{UniForm Network Architectures}
The UniForm Network is constructed around the concept of Reuse Attention, with its architecture illustrated in Fig.~\ref{fig:architecture}. Similar to previous Like previous hierarchical backbones\cite{LiuICCV21Swin, LiuCVPR23EfficientViT, WangICCV21PVT, HowardICCV19MobileNetV3, GrahamICCV21LeViT}, UniForm follows a progressive design that operates in three stages. Across these stages, the channel dimensions $C$, the depth $L$, and the number of attention heads $H$ are incrementally increased to accommodate different levels of feature abstraction.

In the first stage, we introduce overlapping patch embedding to transform 16$\times$16 input patches into tokens of dimension $C_1$. This method enhances the model's capacity for low-level visual representation learning, capturing finer details with minimal computational overhead. The UniForm Network then builds upon these tokens through a series of UniForm Blocks, stacking them within ach stage while reducing the token count by a factor of 4 via downsampling layers. The downsampling in resolution is inspired by the efficiency of hierarchical architectures which maintain spatial relationships while progressively reducing computational complexity~\cite{LiuICCV21Swin, LiuCVPR23EfficientViT}

We employ depthwise convolution (DWConv) and feedforward network (FFN) layers sequentially before and after attention modules to balance local feature extraction and global context understanding efficiently. This combination reduces computational complexity while capturing both low-level and high-level representations. The use of DWConv and FFN layers around attention is a proven technique from prior researches, enhancing model performance by optimizing the flow of information without the high cost of full self-attention~\cite{LiuCVPR23EfficientViT, LiICCV23EfficientFormer, ChenCVPR22MobileFormer}. The architecture is highly scalable, supporting Tiny, Small, Medium, and Large variants (as shown in Table~\ref{tab:model_architecture}), with each variant adjusting the number of channels, attention heads, and depth to meet varying task complexities and computational constraints. UniForm also introduces Reuse Attention blocks, which reuse intermediate features across stages, reducing computational costs without sacrificing accuracy. This modular design enhances flexibility, allowing the network to adapt seamlessly to different patch sizes and resolutions.

\section{Experimental Results}
\label{sec:experiment}
\subsection{Implementation Details}
We conduct image classification experiments on ImageNet-1K~\cite{DengCVPR09ImageNet}. The models are buit with PyTorch 2.3.0~\cite{PaszkeNeurIPS19Pytorch} and MMPreTrain 1.2.0~\cite{2023mmpretrain} , and  trained from scratch for 300 epochs on 2 NVIDIA A100 GPUs using AdamW optimizer~\cite{LoshchilovICLR19AdamW} and cosine annealing learning rate scheduler~\cite{LoshchilovICLR17cosine}. We set the total batchsize as 512. The input images are resized and randomly cropped into 224$\times$224. The initial learning rate is 0.001 with weight decay of 0.025. We include some augmentation and regularization strategies in training, including Mixup~\cite{ZhangICLR18MixUp}, Cutmix~\cite{YunICCV19CutMix}, and 
random erasing~\cite{ZhongAAAI20erasing}.

Additionally, we evaluate the throughput across different hardware plateforms.
\begin{itemize}
    \item For GPU, throughput is measured on an NVIDIA A100, with a batch size of 2048 for a fair comparison across models.
    \item For CPU, we measure the runtime on an Intel Xeon Gold 6426Y @ 2.50 GHz processor using a batch size of 16 and single-thread execution following the methodology based on~\cite{GrahamICCV21LeViT}.
    \item In contrast to prior works, we also extensively emphasize the evaluation on the inference performance on various edge devices. These include popular multiple versions of the \textit{Jetson} (\textit{Nano, Xavier, Tx2, Nx, and AGX Orin}) and multiple versions of the \textit{Raspberry Pi} (\textit{2B, 3B, 3B Plus, 4B, and 5}). All models are tested with a batch size of 16 and run in single-thread mode to maintain consistency. This evaluation demonstrates the practicality of each model in edge computing environments, where resource constraints are significantly more stringent than on server-grade hardware.
\end{itemize}

Moreover, we evaluate the transferability of the UniForm model to downstream tasks. For image classification, we fine-tune the models for 300 epochs following~\cite{ZhangCVPR22MiniViT} with similar hyperparameter settings. For instance segmentation on the COCO dataset~\cite{LinECCV14CoCo}, we use Mask RCNN and train for 12 epochs (1$\times$ schedule) with the same settings as~\cite{LiuICCV21Swin} using the MMDetection framework~\cite{mmdetection}.

\subsection{Image Classification}
\subsubsection{Performance on High-Performance Hardware (GPU/CPU)}
The UniForm models (Tiny, Small, Medium, and Large) consistently outperform state-of-the-art (SOTA) models across a range of sizes, delivering both higher accuracy and superior throughput in Fig.~\ref{fig:throughput} and Tab.~\ref{tab:perform_comparison_acc}. When compared to traditional CNN-based models like ShuffleNetV2 and MobileNetV3, as well as modern Transformer-based models like EfficientViT and EdgeNeXt, UniForm-s achieves 70.1\% Top-1 accuracy, significantly outperforming MobileNetV3-small (67.4\%), EfficientViT-M1 (68.4\%), MobileViT-XXS (69.0\%), and ShuffleNetV2 1.0x (69.4\%) while also offering a higher throughput on CPU (231 images/s) as well as GPU (50,582 images/s vs. 41,965 images/s for MobileNetV3-small and 47,045 images/s for EfficientViT-M1). Furthermore, UniForm outperforms models from other architecture families, including MLP-based models like Mixer-B/16 and fusion-based models like EdgeNeXt. UniForm-l, for example, achieves 76.7\% accuracy with a throughput of 25,356 images/s on GPU, significantly faster than Mixer-L/16 (688 images/s) and ViG-Ti (1,406 images/s), all while delivering higher accuracy than both.

This trend is evident across all UniForm variants, demonstrating that UniForm not only provides better accuracy but also achieves faster throughput on both GPU and CPU compared to other models of similar sizes, making it highly efficient for both large-scale and edge-device environments.

\subsubsection{Inference Speed and Efficiency on Edge Devices}
The results presented in Table \ref{tab:perform_comparison_edgedevice} showcase the inference speed of UniForm variants compared to state-of-the-art CNN and ViT models across a wide range of edge devices. The key takeaway from these findings is UniForm's exceptional performance, consistently achieving faster inference times while maintaining competitive accuracy across different edge-devices.

Across all edge devices, UniForm significantly outperforms its counterparts in terms of inference speed. For example, UniForm-t demonstrates a 5x improvement in speed on the Jetson-Nano (11.9ms) compared to EfficientViT-M0 (56.8ms) while also providing a higher Top-1 accuracy (66.0\% vs. 63.2\%). This trend continues with UniForm-s and UniForm-m, both showing notable improvements in inference times across all Raspberry Pi versions and Jetson devices when compared to models like MobileNetV3 and EdgeNeXt. On the RaspberryPi4B, UniForm-t (19.4ms) is significantly faster than MobileNetV3-small (33.3ms) and EfficientViT-M1 (400.3ms). This highlights its versatility for deployment in resource-constrained environments where power efficiency and speed are critical, providing a viable solution for deploying advanced vision models in real-world edge scenarios..

UniForm excels at balancing accuracy and speed on edge devices, outperforming both CNN-based models like MobileNetV3 and ShuffleNetV2, as well as Transformer-based models like EfficientViT and DeiT-Tiny. For instance, UniForm-l achieves the highest accuracy (76.7\%) with significantly faster inference times on devices like the Jetson AGX Orin (2.4ms) compared to other Transformer models.

\subsection{Downstream Tasks}
We validated the effectiveness of UniForm models on several downstream tasks, focusing on image classification and instance segmentation to showcase the model's adaptability and competitive edge over state-of-the-art architectures.

\subsubsection{Image Classification Downstream Tasks}
We evaluate UniForm across several image classification benchmarks, including CIFAR-10, CIFAR-100, Flowers-102, and Oxford-IIIT Pet. UniForm consistently demonstrates competitive top-1 accuracy across these datasets, maintaining a balance between inference speed and performance. The results show that UniForm efficiently handles different dataset scales, particularly excelling in datasets like Flowers-102 and Oxford-IIIT Pet. This suggests that UniForm transfers well to smaller, fine-grained classification tasks while preserving throughput on both edge devices and traditional GPUs.

\subsubsection{Instance Segmentation}
For instance segmentation, UniForm is tested on object detection tasks using the COCO dataset. When paired with Mask R-CNN, UniForm demonstrates robust performance, achieving competitive segmentation results while maintaining efficient inference on edge devices. This evaluation is essential in environments where high-resolution dense prediction tasks must operate under constrained resources.

\begin{itemize}
    \item Inference Speed Improvement: Achieved up to a 30\% reduction in inference time on edge devices like Raspberry Pi and Jetson Nano compared to the traditional attention mechanism.
    \item Memory Usage Reduction: Observed a substantial decrease in memory consumption, enabling deployment on devices with stringent memory constraints.
    \item Maintained Accuracy: Demonstrated comparable performance on benchmark datasets such as ImageNet, with negligible loss in Top-1 accuracy.
    \item Generalization to Downstream Tasks: Successfully applied our method to tasks like object detection and segmentation, achieving performance gains without additional modifications.
\end{itemize}

\begin{table*}[]
    \centering
    \caption{Inference speed and accuracy comparison between UniForm and state-of-the-art CNN and ViT models across various edge devices.}
    \label{tab:perform_comparison_edgedevice}
    \scalebox{0.8}{
    \setlength{\tabcolsep}{6pt}
\begin{tabular}{ccccccccccccc}
\toprule
Model                 & \rotatebox[origin=c]{60}{Top-1 Accuracy} & \rotatebox[origin=c]{60}{Jetson-Nano}   & \rotatebox[origin=c]{60}{RaspberryPi4B} & \rotatebox[origin=c]{60}{RaspberryPi3B Plus} & \rotatebox[origin=c]{60}{RaspberryPi3B}  & \rotatebox[origin=c]{60}{RaspberryPi2B}  & \rotatebox[origin=c]{60}{Jetson-Xavier} & \rotatebox[origin=c]{60}{Jetson-Tx2}    & \rotatebox[origin=c]{60}{Jetson-Nx}    & \rotatebox[origin=c]{60}{RaspberryPi5}  & \rotatebox[origin=c]{60}{Jetson AGX Orin} \\
\midrule
EfficientViT-M0       & 63.2           & 56.8          & 247.6         & 521.1              & 528.8          & 1908.3         & 10.5          & 25.3          & 14.1         & 93.3          & 6.0             \\
\textbf{UniForm-t}    & \textbf{66.0}  & \textbf{11.9} & \textbf{19.4} & \textbf{35.7}      & \textbf{40.5}  & \textbf{141.7} & \textbf{4.4}  & \textbf{7.7}  & \textbf{5.8} & \textbf{6.4}  & \textbf{2.0}    \\
\midrule
MobileNetV3-small     & 67.4           & 7.3           & 33.3          & 65.4               & 73.5           & 233.7          & 1.9           & 4.0           & 2.6          & 12.2          & 1.1             \\
EfficientViT-M1       & 68.4           & 74.7          & 400.3         & 839.1              & 863.1          & 3162.2         & 12.0          & 32.3          & 17.8         & 147.7         & 6.7             \\
MobileViT-XXS         & 69.0           & 75.1          & 84.6          & 156.8              & 172.0          & 549.4          & 5.0           & 10.5          & 7.1          & 31.7          & 2.1             \\
ShuffleNetV2 1.0x     & 69.4           & 8.6           & 32.6          & 62.2               & 69.6           & 224.9          & 2.1           & 4.2           & 3.0          & 9.7           & 1.1             \\
\textbf{UniForm-s}    & \textbf{70.1}  & \textbf{13.5} & \textbf{31.9} & \textbf{60.4}      & \textbf{67.0}  & \textbf{244.0} & \textbf{4.4}  & \textbf{8.0}  & \textbf{6.0} & \textbf{9.7}  & \textbf{2.0}    \\
\midrule
EfficientViT-M2       & 70.8           & 84.9          & 473.1         & 975.2              & 985.6          & 3756.6         & 13.3          & 36.4          & 20.3         & 158.2         & 7.1             \\
EdgeNeXt-XXS          & 71.2           & 64.6          & 59.4          & 105.3              & 120.0          & 390.7          & 4.1           & 8.5           & 5.7          & 22.1          & 2.0             \\
MobileOne-S0          & 71.4           & 8.9           & 54.2          & 108.7              & 122.7          & 427.5          & 1.7           & 3.7           & 2.7          & 16.9          & 0.8             \\
Mixer-B/16            & 71.7           & 991.6         & 987.8         & N/A                & N/A            & N/A            & 32.8          & 80.0          & 50.2         & 321.1         & 5.8             \\
RepVGG-A0             & 72.4           & 15.0          & 125.0         & 249.8              & 284.8          & 1205.0         & 2.8           & 6.1           & 3.8          & 33.9          & 0.9             \\
EfficientViT-M3       & 73.4           & 101.6         & 568.1         & 1136.1             & 1223.9         & 4764.8         & 16.5          & 43.4          & 24.7         & 187.0         & 8.0             \\
ViG-Ti                & 73.9           & 76.0          & 214.4         & 391.8              & 430.5          & 1943.5         & 14.0          & 32.7          & 21.1         & 98.4          & 5.3             \\
\textbf{UniForm-m}    & \textbf{74.1}  & \textbf{15.8} & \textbf{43.7} & \textbf{81.9}      & \textbf{93.3}  & \textbf{346.7} & \textbf{4.3}  & \textbf{9.9}  & \textbf{6.7} & \textbf{13.7} & \textbf{2.0}    \\
\midrule
EfficientViT-M4       & 74.3           & 108.6         & 634.9         & 1264.7             & 1352.8         & 5263.7         & 17.2          & 45.6          & 25.9         & 201.7         & 8.3             \\
RepVGG-A1             & 74.4           & 22.4          & 212.4         & 425.6              & 478.4          & 1991.4         & 3.6           & 9.0           & 6.0          & 73.2          & 1.1             \\
DeiT-Tiny (distilled) & 74.5           & 39.0          & 134.0         & 237.5              & 260.1          & 1109.6         & 7.5           & 16.6          & 10.2         & 49.9          & 2.0             \\
MobileViT-XS          & 74.7           & 36.4          & 196.1         & 353.7              & 389.0          & 1290.3         & 7.0           & 16.3          & 10.3         & 84.3          & 2.9             \\
EdgeNeXt-XS           & 75.0           & 115.0         & 109.9         & 186.2              & 209.9          & 723.9          & 6.0           & 13.1          & 8.3          & 42.3          & 2.8             \\
RepVGG-B0             & 75.1           & 264.4         & 269.7         & 546.3              & 608.0          & 2529.4         & 4.4           & 11.0          & 7.3          & 91.1          & 1.3             \\
MobileNetV3-large     & 75.2           & 14.9          & 84.2          & 163.5              & 181.7          & 566.5          & 3.1           & 6.7           & 4.2          & 29.1          & 1.7             \\
MobileOne-S1          & 75.9           & 119.0         & 130.1         & 228.7              & 265.3          & 973.6          & 3.2           & 7.5           & 4.9          & 35.8          & 1.2             \\
ConvNeXtV2-Atto       & 76.2           & 146.4         & 156.1         & 294.5              & 318.2          & 993.1          & 6.6           & 14.5          & 8.9          & 62.8          & 2.7             \\
Mixer-L/16            & 76.4           & N/A           & 4113.5        & N/A                & N/A            & N/A            & 105.6         & N/A           & N/A          & 1323.3        & 14.2            \\
RepVGG-A2             & 76.4           & 43.6          & 416.5         & 885.7              & 960.8          & 4170.4         & 6.9           & 16.8          & 10.6         & 145.3         & 1.8             \\
\textbf{UniForm-l}    & \textbf{76.7}  & \textbf{19.6} & \textbf{69.7} & \textbf{134.3}    & \textbf{148.7} & \textbf{567.5} & \textbf{5.1}  & \textbf{12.0} & \textbf{7.8} & \textbf{21.8} & \textbf{2.4}   \\
 \bottomrule
\hline
\end{tabular}
}
\end{table*}

\begin{figure*}[h]
\centering
\includegraphics[width=1.0\linewidth]{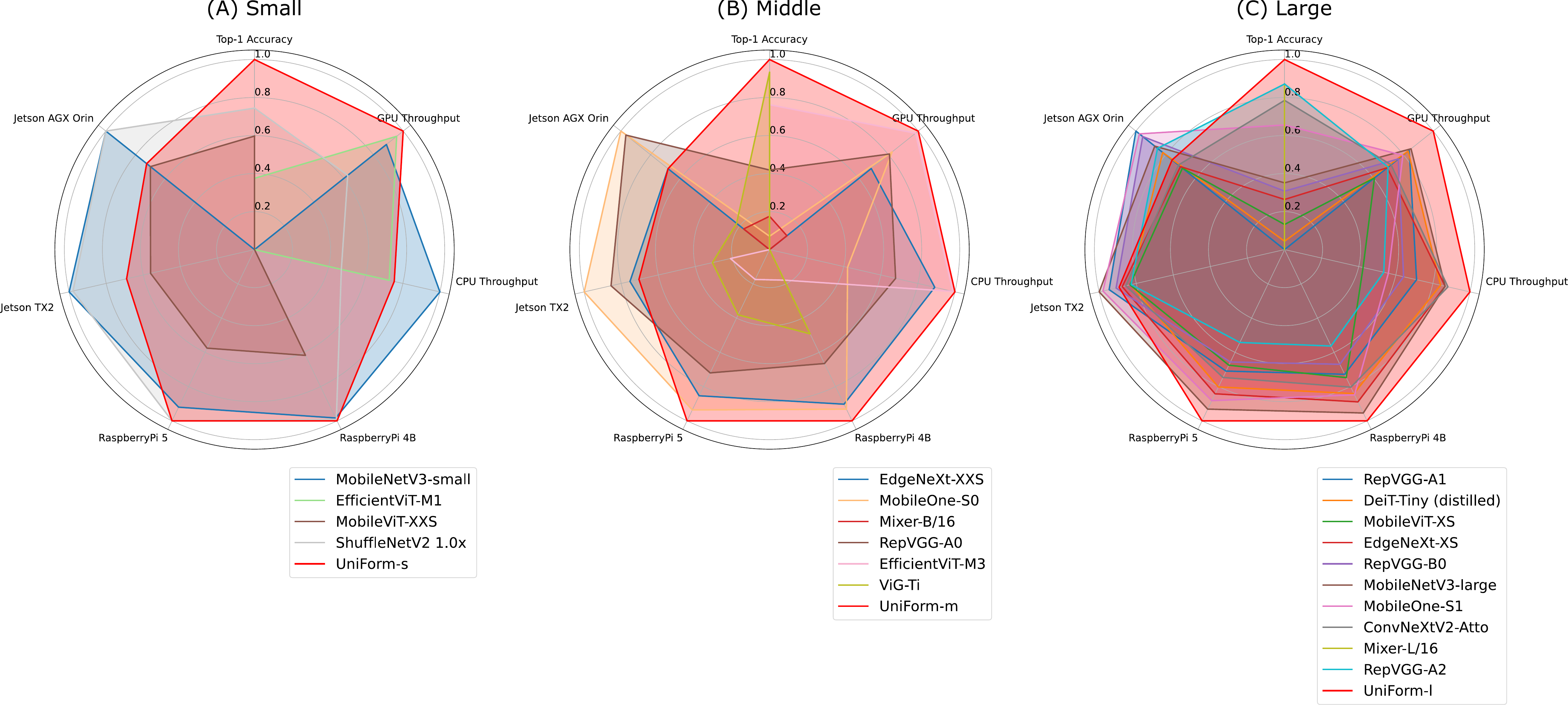}
\caption{Chart comparison of UniForm models with comparable state-of-the-art models across different sizes on a variety of metrics (Top-1 Accuracy and GPU/CPU/Edge-device throughput).}
\label{fig:comparison_feature_map}
\end{figure*}

\begin{figure*}[h]
\centering
\includegraphics[width=0.7\linewidth]{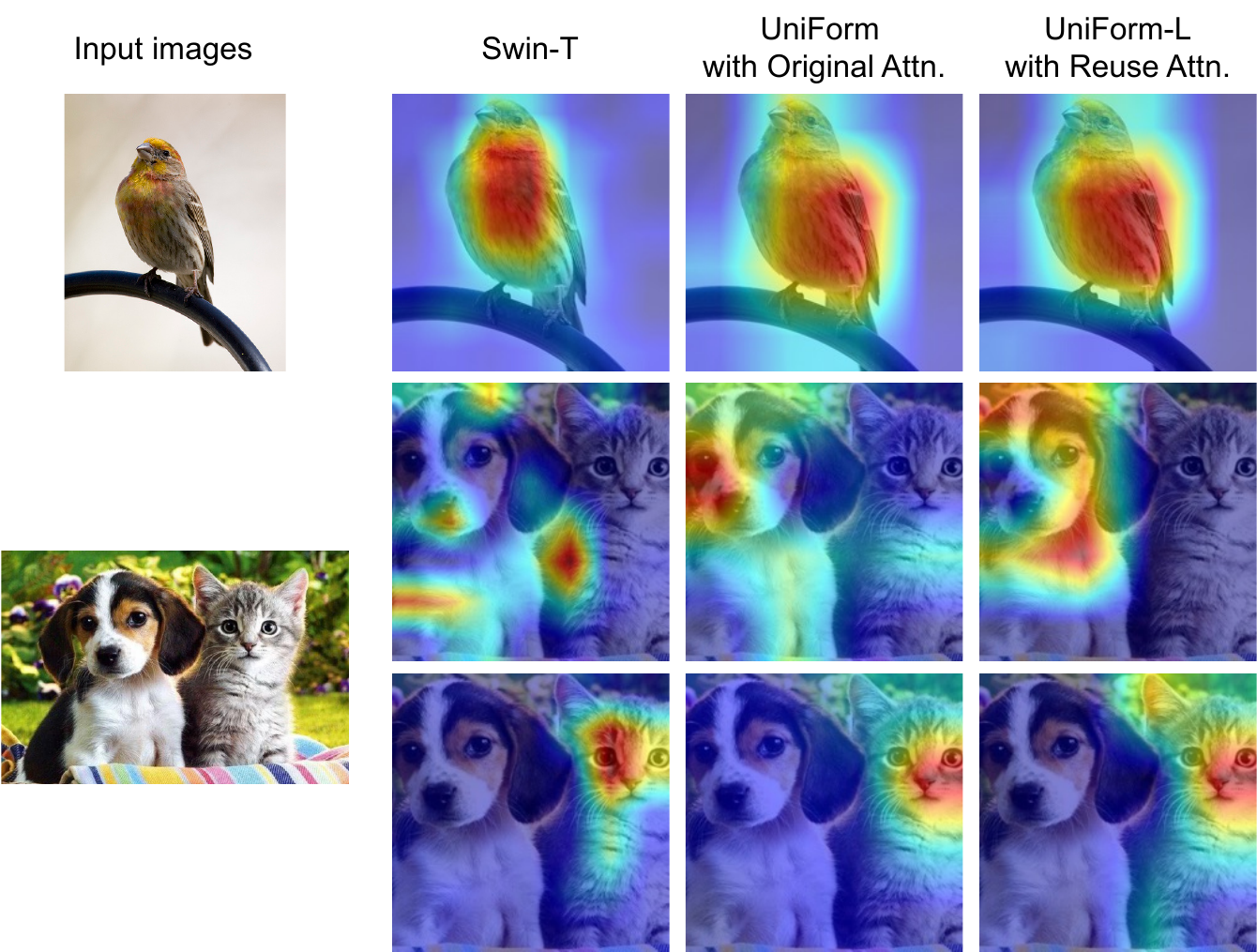}
\caption{Visualization of feature maps prodyced by three different models (Swin-T, UniForm without reuse attention, and UniForm with reuse attention) in the last building block. We use Grad-CAM as our visualization tool. It is obvious that our method with reuse attention can more precisely locate the objects of interest than other methods}
\label{fig:heatmap}
\end{figure*}

\begin{table}[]
    \centering
    \caption{Results of UniForm and other state-of-the-art models on various downstream image classification datasets (CIFAR-10, CIFAR-100, Flowers-102, and Oxford-IIIT Pet).}
    \label{tab:perform_comparison_downstream}
    \newcommand{\spheading}[3][6em]{
    \rotatebox{90}{\parbox{#1}{\raggedright #2}}}
    \scalebox{0.69}{
    \setlength{\tabcolsep}{6pt}
    \begin{tabular}{c|cc|c|c|cccc}
    \toprule
    \multirow{6}{*}{Model} & \multicolumn{2}{c|}{\multirow{3}{*}{Throughput~$\uparrow$}} & \multirow{6}{*}{\spheading{Inference time on Edge ~$\downarrow$}} & \multirow{3}{*}{\rotatebox[origin=c]{90}{ImageNet}} & \multirow{3}{*}{\rotatebox[origin=c]{90}{CIFAR-10}} & \multirow{3}{*}{\rotatebox[origin=c]{90}{CIFAR-100}} & \multirow{3}{*}{\rotatebox[origin=c]{90}{Flowers-102}} & \multirow{3}{*}{\rotatebox[origin=c]{90}{Oxford-IIIT Pet}} \\
                          &  &  &  & & & & & \\
                          & & &  & & & & & \\
                          & \multirow{3}{*}{GPU} & \multirow{3}{*}{CPU} &  & & & & & \\
                        & & &  & & & & & \\
                            & & &  & & & & & \\
    \midrule
    DeiT-Tiny             & 13785           & 63           & 49.9           & 74.5                      & 98.1                      & 86.3                       & 96.9                         & 91.5                             \\
    MobileViT-XS          & 6098            & 13           & 84.3           & 74.7                      & 97.5                      & 84.1                       & 96.1                         & 91.9                             \\
    RepVGG-B0             & 10868           & 30           & 91.1           & 75.1                      & 97.7                      & 85.4                       & 96.0                         & 91.2                             \\
    ConvNeXtV2-Atto       & 9120            & 73           & 62.8           & 76.2                      & 97.2                      & 83.4                       & 87.7                         & 71.4                             \\
    MobileOne-S1          & 12150           & 22           & 35.8           & 75.9                      & 97.7                      & 85.8                       & 97.4                         & 92.2                             \\
    RepVGG-A2             & 8483            & 20           & 145.3          & 76.4                      & 97.9                      & 85.5                       & 95.9                         & 91.9                             \\
    EfficientViT-M5       & 21572           & 101          & 326.5          & 77.1                      & 98.0                      & 86.4                       & 97.1                         & 92.0                             \\
    DeiT-small            & 5768            & 17           & 147.4          & \textbf{80.6}             & \textbf{98.5}             & \textbf{87.2}              & 95.7                         & 91.8                             \\
    \textbf{UniForm-l}    & \textbf{25356}  & \textbf{113} & \textbf{21.8}  & 76.7                      & 98.2                      & 86.5                       & \textbf{97.5}                & \textbf{92.2}                   \\
    \bottomrule
    \end{tabular}
    }
\end{table}

\begin{table}[]
    \centering
    \caption{Performance comparison of instance segmentation on COCO2017}
    \label{tab:perform_comparison_downstream}
    \newcommand{\spheading}[3][6em]{
    \rotatebox{90}{\parbox{#1}{\raggedright #2}}}
    \scalebox{0.69}{
    \setlength{\tabcolsep}{6pt}
\begin{tabular}{ccccccc}
\toprule
Model             & $\text{AP}^{b}$ &  $\text{AP}^{b}_{50}$     & $\text{AP}^{b}_{75}$     & $\text{AP}^{m}$           & $\text{AP}^{m}_{50}$         & $\text{AP}^{m}_{75}$           \\
\midrule
MobileNetV2       & 29.6                  & 48.3          & 31.5          & 27.2          & 45.2          & 28.6          \\
MobileNetV3       & 29.2                  & 48.6          & 30.3          & 27.1          & 45.5          & 28.2          \\
FairNAS-C         & 31.8                  & 51.2          & 33.8          & 29.4          & 48.3          & 31.0          \\
EfficientNet-B0   & 31.9                  & 51.0          & 34.5          & 29.4          & 47.9          & 31.2          \\
MNASNet-A1        & 32.1                  & 51.9          & 34.2          & 29.7          & 49.0          & 31.4          \\
EfficientViT-m4   & 32.8                  & 54.4          & 34.5          & 31.0          & 51.2          & 32.2          \\
\textbf{UniForm-l} & \textbf{33.2}         & \textbf{54.9} & \textbf{35.3} & \textbf{31.5} & \textbf{51.8} & \textbf{32.8} \\
\bottomrule
\end{tabular}
    }
\end{table}

\subsection{Ablation study}
In this section, we ablate ~~~

\textit{Impact of Reuse Attention on Interpretability and Inference Efficiency}
We compare the proposed UniForm model with Reuse Attention against Swin-T and UniForm without Reuse Attention (i.e., with standard attention). As shown in Fig.~\ref{fig:heatmap}, the CAM visualizations demonstrate that UniForm with Reuse Attention preserves strong interpretability, effectively highlighting relevant regions, similar to Swin-T and the UniForm variant without Reuse Attention. Despite the architectural simplicity of UniForm with Reuse Attention, it maintains comparable interpretability while significantly improving inference time, making it more efficient for real-time applications. This showcases the advantage of Reuse Attention, balancing between interpretability and computational efficiency.
\section{Conclusions}
\label{sec:conclusion}
Our Reuse Attention Mechanism offers a simple yet effective solution to the computational and memory inefficiencies of the traditional attention mechanism in Transformers. By leveraging the redundancy in Query and Key components and minimizing memory access operations, we address the critical limitations that hinder the deployment of ViTs on edge devices. This approach not only reduces computational overhead but also aligns with the need for optimizing memory access patterns, directly impacting inference speed and efficiency. Our method demonstrates that significant performance improvements can be achieved through straightforward modifications, paving the way for more practical applications of Transformers in resource-constrained environments.
{
    \small
    \bibliographystyle{ieeenat_fullname}
    \bibliography{main}
}


\end{document}